\documentclass[lettersize,journal]{IEEEtran}
\usepackage{amsmath,amsfonts}
\usepackage{algorithmic}
\usepackage{algorithm}
\usepackage{array}
\usepackage[caption=false,font=normalsize,labelfont=sf,textfont=sf]{subfig}
\usepackage{textcomp}
\usepackage{stfloats}
\usepackage{url}
\usepackage{verbatim}
\usepackage{float}
\usepackage{graphicx}
\usepackage{multirow}
\usepackage{cite}
\usepackage{color}
\begin{document}

\title{Multi-query Vehicle Re-identification: Viewpoint-conditioned Network,
Unified Dataset and New Metric}

\author{
		Aihua Zheng,
		Chaobin Zhang,
		Weijun Zhang,
		Chenglong Li*,
		Jin Tang,
		Chang Tan,
		Ruoran Jia,

	    \thanks{This research is supported in part by the National Natural Science Foundation of China (61976002), the University Synergy Innovation Program of Anhui Province (GXXT-2020-051 and GXXT-2019-025), and the Natural Science Foundation of Anhui Province (2208085J18).}
		    
		\thanks{
		A. Zheng and C. Li are with the Information Materials and Intelligent Sensing Laboratory of Anhui Province, Anhui Provincial Key Laboratory of Multimodal Cognitive Computation, School of Artificial Intelligence, Anhui University, Hefei, 230601, China
		(e-mail: ahzheng214@foxmail.com; lcl1314@foxmail.com).}
			
		\thanks{C. Zhang, W. Zhang and J. Tang are with Anhui Provincial Key Laboratory of Multimodal Cognitive Computation, School of Computer Science and Technology, Anhui University, Hefei, 230601, China (e-mail:  chaobinzhang@foxmail.com; zwj\_edu@foxmail.com;  tangjin@ahu.edu.cn)}
		
		\thanks{
		C. Tan and R. Jia are with iFLYTEK Co., Ltd., Hefei 230088, China.
        (Email: changtan2@iflytek.com, rrjia@iflytek.com ).
		}
		 
	}


\maketitle

\begin{abstract}
Existing vehicle re-identification methods mainly rely on the single query, which has limited information for vehicle representation and thus significantly hinders the performance of vehicle Re-ID in complicated surveillance networks.
In this paper, we propose {\bf a more realistic and easily accessible task}, called multi-query vehicle Re-ID, which leverages multiple queries to overcome viewpoint limitation of single one. 
Based on this task, we make three major contributions.
First, we design \emph{a novel viewpoint-conditioned network} (VCNet), which adaptively combines the complementary information from different vehicle viewpoints, for multi-query vehicle Re-ID. 
Moreover, to deal with the problem of missing vehicle viewpoints, we propose a cross-view feature recovery module which recovers the features of the missing viewpoints by learnt the correlation between the features of available and missing viewpoints.
Second, we create \emph{a unified benchmark dataset}, taken by 6142 cameras from a real-life transportation surveillance system, with comprehensive viewpoints and large number of crossed scenes of each vehicle for multi-query vehicle Re-ID evaluation.
Finally, we design \emph{a new evaluation metric}, called mean cross-scene precision (mCSP), which measures the ability of cross-scene recognition by suppressing the positive samples with similar viewpoints from same camera.
Comprehensive experiments validate the superiority of the proposed method against other methods, as well as the effectiveness of the designed metric in the evaluation of multi-query vehicle Re-ID.
\end{abstract}


\section{Introduction}
\IEEEPARstart{V}{ehicle} re-identification (Re-ID) aims to correlate the images of the same vehicle captured by non-overlapping cameras. 
This task has been widely applied in urban security monitoring and intelligent transportation systems, and has received more and more attention in recent years~\cite{TIP19Louvehicle,TIP21Hsuvehicle,TIP21Dilshadvehicle,TIP21Guovehicle,Liu19Adaptive,luo2019bag,guo2018learning,vehicle20Mai,vehicle21Dilshad,liu2020beyond,alfasly2019multi,TICS21Shen,TICS22Lu,TPAMI21Bai,TMM21Li}.
\begin{figure}[H]
\begin{center}
\includegraphics[width=\linewidth]{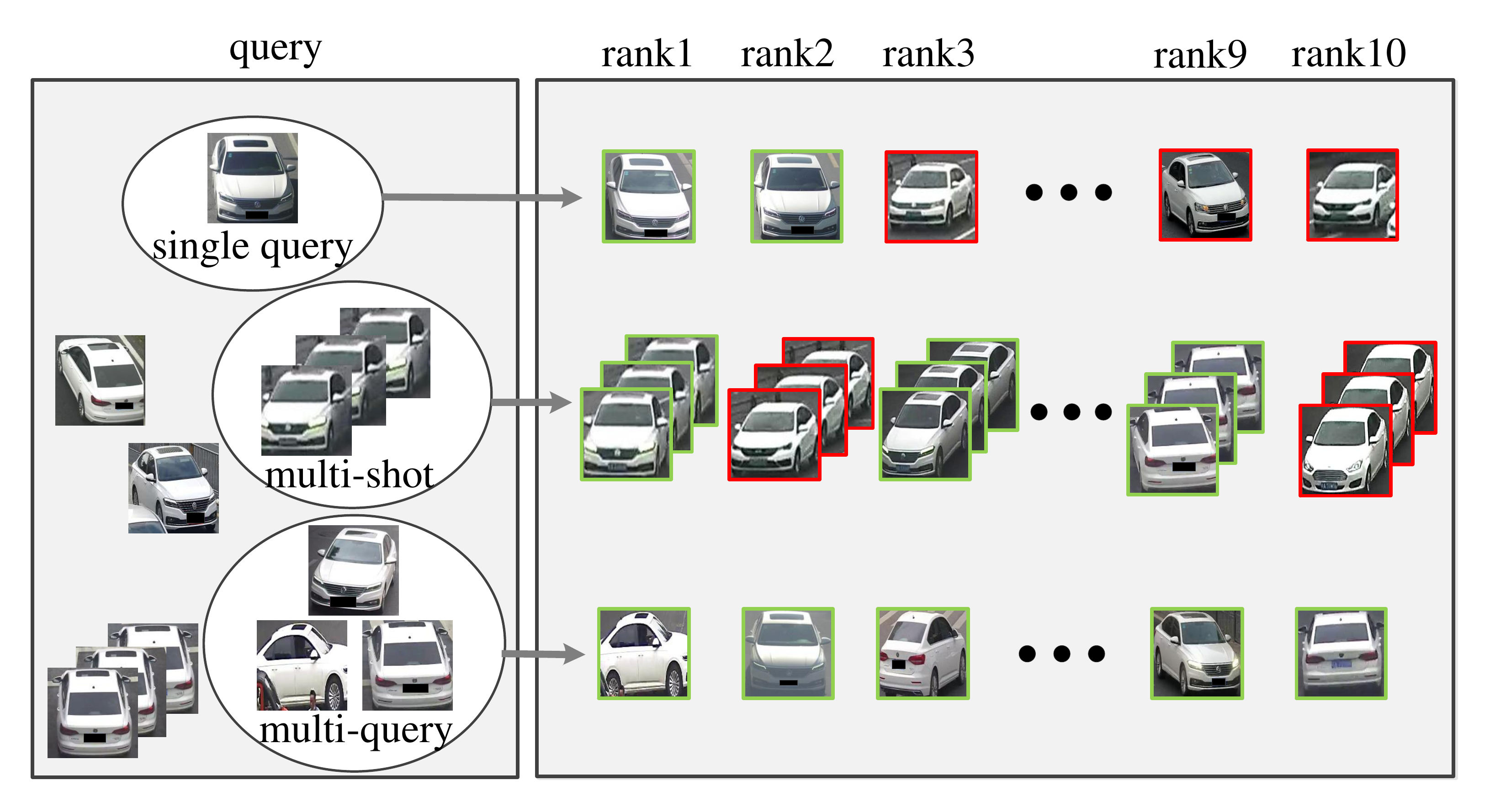}
\end{center}
\caption{
Examples of ranking results comparison among the conventional single query, multi-shot and the proposed multi-query ReID on our collected MuRI dataset via ResNet-50~\cite{he2016deep}, where multi-shot (or multi-query) ReID is achieved by averaging the consecutive (or the different viewpoints) vehicle image features. The true and false matchings are bounded in green and red boxes, respectively. 
}
\label{fig:motivation}
\end{figure}

Most existing image-based vehicle Re-ID methods~\cite{chen2019dmml,zhu20rectnet,liu20grf,rao2021cal,chen2020vehicle,he2021transreid} rely on the single query.
However, the dramatic appearance changes caused by different viewpoints lead to the huge intra-class discrepancy.
%
%

To solve the viewpoint diversity, some works use vehicle keypoint information~\cite{wang17keypoint,moskvyak2021keypoint} or vehicle local area features~\cite{Liu18RAM,meng2020parsing,sun2021tbe} to perform local feature alignment.
Moreover, meta-information (e.g. vehicle attributes, spatial-temporal information) has also been explored to alleviate the difference from different viewpoints.
Zheng \textit{et al.}~\cite{zheng19Attributes} introduce attribute fusion and Liu \textit{et al.}~\cite{Liu18PROVID} utilize attributes and spatial-temporal information to learn global vehicle representations.
However, the key issue in this single-shot fashion is that one vehicle image only has a specific viewpoint as the single query, thus it is very challenging to match the gallery images with different viewpoints.
To explore the comprehensive information in multi-shot images, Jin \textit{et al.}~\cite{Jin20multishot} propose a multi-shot teacher branch that mines the information of multi-viewpoint images to guide the single-image student branch during training.
However, they cannot guarantee that the teacher network contains information from multiple viewpoints, and they  only use one query image in the inference phase, which still can not fully utilize the information from multiple query images. 
Zheng \textit{et al.}~\cite{Zheng15AP} evaluate person Re-ID in the so-called multi-query fashion by average or max operations on the multiple person images from the same camera to obtain a new query feature in the inference process. 
However, the variability between query images in the same camera is small, and the average or max operations cannot fully utilize the diversity and complementarity among the multiple queries.

\textbf{In fact, we can easily access multiple images of a certain vehicle from diverse viewpoints or scenes as the query in the real-life surveillance.}
On the one hand, in a certain scene, we can easily obtain the multi-view images of the same vehicle via the crowded dome and box cameras or the robust tracking algorithms.
On the other hand, we can obtain the cross-scene multi-view images of the same vehicle by correlating the corresponding cross-scene tracklets.
In addition to the above intelligent acquisition, we can also manually construct the multi-view query images in the surveillance system.

By integrating the complementary information among the multi-query images scenario, we can learn more comprehensive appearance feature representation of a certain vehicle, which is expected to significantly overcome the dramatic appearance changes caused by different viewpoints.
Therefore, we rethink vehicle Re-ID in the more realistic multi-query inference setting in this paper. 
%
%
As shown in Fig.~\ref{fig:motivation}, due to the limited view information, single query and multi-shot Re-ID tend to easily matching the vehicle images with similar viewpoints. By contrast, multi-query Re-ID can hit the more challenging right matchings with diverse viewpoints since it can integrate the complementary information among the diverse viewpoint queries. 
%
Giving the easily accessible multiple query images captured from a single or several non-overlapping scenes/cameras, how to take the advantage of multiple queries with diverse viewpoint and illumination changes to achieve more accurate vehicle Re-ID?
\textbf{In this paper, we propose a novel viewpoint-conditioned network (VCNet), which effectively combines the complementary information from different vehicle viewpoints, for multi-query vehicle Re-ID.}
First, in the training process, to make full use of diverse viewpoint information of the vehicle, 
we propose a viewpoint conditional coding (VCC) module, which uses the learned vehicle viewpoint features as viewpoint conditional coding information and integrates it into the feature learning process of vehicle appearance.
Second, we propose the viewpoint-based adaptive fusion (VAF) module to adaptively fuse the viewpoint coded appearance features of the vehicle in the multi-query inference process.
In particular, it adaptively assigns weights to the appearance features of the multiple queries according to their viewpoint similarity to gallery.  
The higher viewpoint similarity between the query image to the current gallery image, the larger weight to the appearance feature of the corresponding query image.
\textbf{Finally, to tolerate the missing viewpoints in the query set in the inference, we propose a cross-view feature recovery module (CVFR) to recover the features of the missing viewpoints.}

In addition, although conventional vehicle Re-ID metrics (namely CMC, mAP and mINP) avoid the easy matching from the same camera between query and gallery, they mainly focus on the global relation between the query and gallery while ignoring the local relations within the gallery.
Therefore, they tend to result in virtual high scores when retrieving easy positive samples with similar viewpoints from one single camera.
Although Zhao \textit{et al.}~\cite{Zhao21CGM} propose the evaluation metric Cross-camera Generalization Measure (CGM) to improve the evaluations by introducing position-sensitivity and cross-camera generalization penalties.
It still suffers from the influence of similar viewpoint samples under the same camera. 
Since they only easily divide target images captured from the same cameras into individual groups, and fail to consider the positive samples with similar viewpoints from the individual group.
In this paper, we argue that the realistic Re-ID cares more about the cross-scene retrieval ability of the model, which is more crucial to the intelligent transportation society to trace the trajectory of the certain vehicle among the identity of the vast Skynet in the smart city.
\textbf{Therefore, we propose a new metric, the mean Cross-scene Precision (mCSP), which focuses on the cross-scene retrieval ability by suppressing the positive samples with similar viewpoints from the same camera.}

At last, although existing vehicle Re-ID datasets, including VehicleID~\cite{Liu16CMC}, VeRI-776~\cite{Liu16veri776}, and VERI-Wild ~\cite{Lou19veriwild}, provide important benchmarks to evaluate the state-of-the-art methods, the crucial issue is the number of cameras is limited (12 in Vechile ID, 20 in VeRI-776 and 174 in VERI-Wild).
Therefore, each vehicle only appears with limited cameras.
Furthermore, although they contain vehicle images from multiple viewpoints, the number of viewpoints for each vehicle ID is still limited.
Herein,
\textbf{we propose a new vehicle image dataset captured by large amount of cameras (\textit{i.e.}, 6142 cameras) from a
real-life transportation surveillance system, named Multi-query Re-Identification dataset (MuRI). }
MuRI contains diverse viewpoints, including $front$ ($side$ $front$), $side$ and $rear$ ($side$ $rear$), and large number of crossed scenes/cameras for each vehicle (\textit{i.e.}, 34.6 in average), which provides more realistic and challenging scenarios for multi-query vehicle Re-ID.

To the best of our knowledge, we are the first to launch the multi-query setting in vehicle Re-ID, which jointly uses images from multiple scenes/viewpoints of a vehicle as a query.
The contributions of this paper are mainly in the following four aspects.
\begin{itemize}

\item We introduce a new task called multi-query vehicle re-identification, which devotes to inferring the cross-scene re-identification by exploring the complementary information among the  multiple query images with different viewpoints. The task is challenging, but easily accessible and very practical in realistic transportation systems.
%

\item We propose a viewpoint-conditioned network (VCNet) for multi-query vehicle Re-ID, which learns special viewpoint information through the viewpoint conditional coding (VCC) module during the training and integrate the complementary viewpoints information through the viewpoint-based adaptive fusion (VAF) module in the testing. In addition, we propose the cross-view feature recovery (CVFR) module to deal with the missing viewpoint problem.
%
%
%

\item To measure the cross-scene retrieval ability of Re-ID, we further design a new metric, namely mean cross-scene precision (mCSP), by suppressing the positive samples with similar viewpoints from the same camera, which provides a more crucial measure in real-life Re-ID applications.
\item To evaluate the effectiveness of the proposed VCNet for multi-query inference in vehicle Re-ID, we collect a multi-query vehicle Re-ID dataset with a large number of crossed scenes  from the real city traffic.

\end{itemize}

\section{Related Work}
\subsection{Vehicle Re-ID Methods}
Most of the vehicle Re-ID methods rely on single query image. In order to learn the detailed features of vehicles and expand the subtle differences between the same models, some works introduce the idea of region of interest prediction or attention models to mine the salient regions of vehicles. 
He \textit{et al.}~\cite{he2019part} propose a simple and effective partial regularization method, which detects the regions of interest using pre-trained detectors and introduce multi-dimensional constraints at the part level (windows, lights, and make alike) into the vehicle Re-ID framework. It improves the model's ability to learn local information while enhancing the subtle difference perception. 
Zhang \textit{et al.}~\cite{an2019part} propose an attention network using local region guidance, which mines the most important local regions by learning the weights of candidate search regions to increase the weights of discriminative features in vehicle images while reducing the effect of irrelevant background noise.
Khorramshshi \textit{et al.}~\cite{Khorramshshi19A} learn to capture localized discriminative features by focusing attention on the most informative key-points based on different orientation.

To handle the similar appearance of the different vehicles, some works propose to use the additional annotation information of the dataset to learn more accurate local features of the vehicle.
Liu \textit{et al.}~\cite{Liu18PROVID} exploit multi-modal data from large-scale video surveillance, such as visual features, license plates, camera locations, and contextual information, to perform coarse-to-fine search in the feature domain and near-to-far search in the physical space.
%
Wang \textit{et al.}~\cite{wang17keypoint} extract local area features in different directions based on 20 keypoint locations, which are aligned and combined by embedding into directional features. The spatio-temporal constraints are modeled by spatio-temporal regularization using log-normal distribution to refine the retrieval results.
%
%
Zheng \textit{et al.}~\cite{zheng19Attributes} introduce a deep network to fuse the camera views, vehicle types and color into the vehicle features.

Metric learning based approaches focus on solving the problem of intra-class variation and inter-class similarity caused by view variation. 
Bai \textit{et al.}~\cite{Yan18Group} propose a deep metric learning method that divides samples within each vehicle ID into groups using an online grouping method, and create multi-granularity triple samples across different vehicle IDs as well as different groups within the same vehicle ID to learn fine-grained features. 
Jin \textit{et al.}~\cite{jin2021model} propose a multi-center metric learning framework for multi-view vehicle Re-ID that models potential views directly from the visual appearance of vehicles, and constrains the vehicle view centers by intra-class ranking loss and cross-class ranking loss to increase the discriminative information of different vehicles.

%
To explore more information in the query, Jin \textit{et al.}~\cite{Jin20multishot} explore the comprehensive information of multi-shot images of an object in a teacher-student manner. 
Although they use the multi-shot teacher branch to guide the single-image branch during training, it still contained only single-image information during the inference phase.

\subsection{Vehicle Re-ID Metrics}
Vehicle Re-ID is an image retrieval subproblem, and to evaluate the performance of Re-ID methods. Cumulative Matching Characteristics (CMC)~\cite{Liu16CMC} and mean Average Precision (mAP)~\cite{Zheng15AP} are two widely used measures.
CMC-k (also known as k-level matching accuracy)~\cite{Liu16CMC} indicates the probability of a correct match among the top k ranked retrieval results. When comparing the performance of different methods, if there is little difference in performance between methods, the cumulative matching performance curves will overlap for the most part, making it impossible to accurately determine good or bad performance. In order to compare the performance differences between methods more concisely, the cumulative matching accuracy at some key matching positions is generally selected for comparison, where rank1 and rank5 are more common, indicating the probability of correctly matching the first 1 and the first 5 images in the result sequence, respectively.

Another metric, the mean accuracy (mAP)~\cite{Zheng15AP}, is used to evaluate the overall performance of the Re-ID methods and represents the average of the accuracy of all retrieval results. It is originally widely used in image retrieval.
For Re-ID evaluation, it can address the issue of two systems
performing equally well in searching the first ground truth, but has different retrieval abilities for other hard matches.
However, these two widely used measures cannot assess the ability of the model to retrieve difficult samples. To address this issue, Ye \textit{et al.}~\cite{ye21AGW} propose a computationally efficient metric, namely a negative penalty (NP), which measures the penalty to find the hardest correct match.
To measure the results derived from individual cameras, \textit{et al.}~\cite{Zhao21CGM} propose a cross-scene generalization measure (CGM). It first divides the vehicle images captured by the same camera into individual groups, then calculate the average ranking values for each camera.

%
However, all of the above metrics ignore the similar positive samples in the same camera, which leads to the virtual high metric scores.

\begin{figure*}[ht]
\begin{center}
\includegraphics[width=0.95\textwidth]{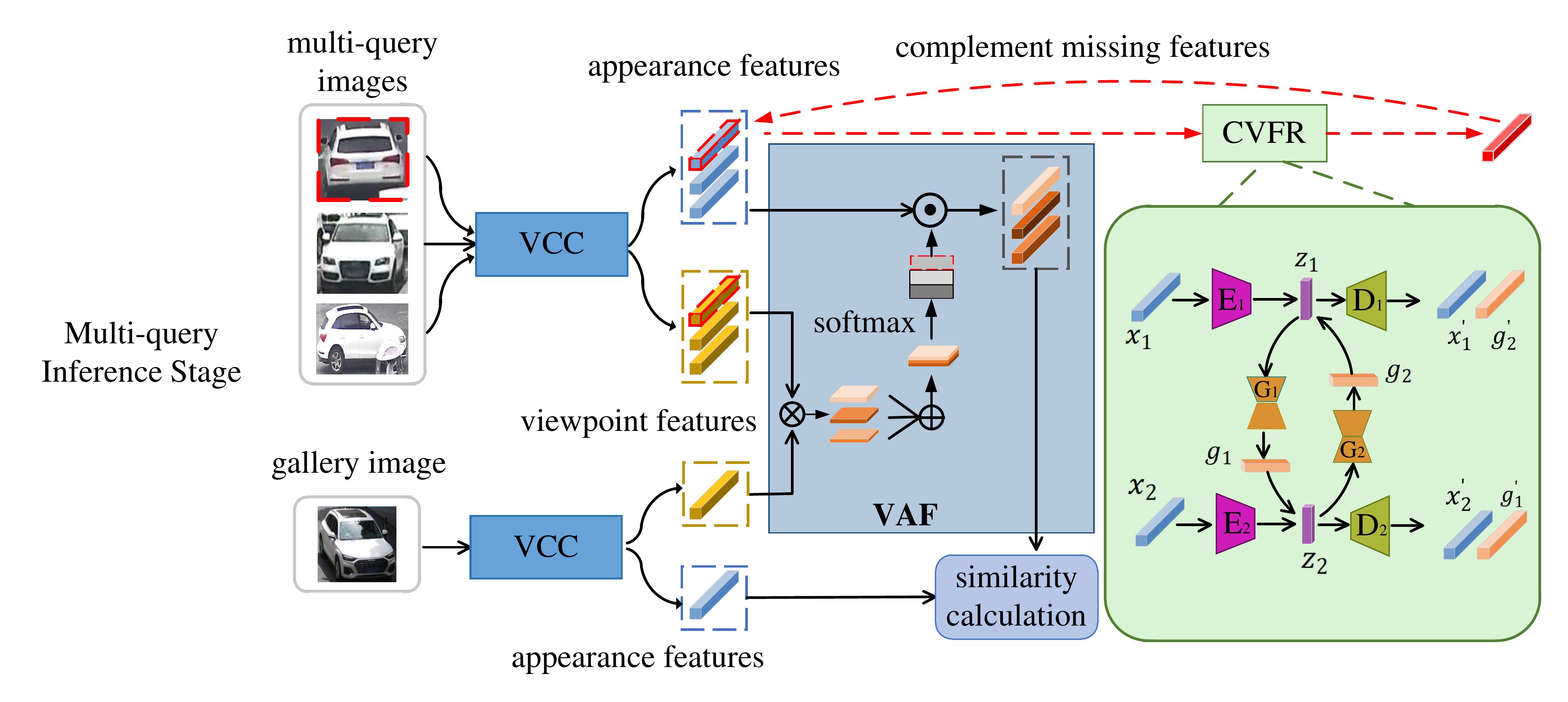} 
\end{center}
\caption{
The pipeline of our inference framework. In the multi-query inference stage, viewpoint weights will be calculated between query and gallery viewpoint features, to integrate the complementary information among different viewpoints of the vehicle, we adaptively fuse the generated viewpoint weights with appearance features by viewpoint-based adaptive fusion (VAF) module. When we miss a query image from a random viewpoint, the appearance feature will be recovered by the cross-view feature recovery (CVFR) module. "$\oplus$", "$\otimes$" and "$\odot$" denote concatenation, cosine similarity calculation, and multiply respectively.
}
\label{fig:test}
\end{figure*}

\subsection{Vehicle Re-ID Datasets}
Recent vehicle Re-ID methods are mainly evaluated on three public datasets, including VeRI-776~\cite{Liu16veri776}, VehicleID~\cite{Liu16CMC} and VERI-Wild~\cite{Lou19veriwild}.
VeRI-776~\cite{Liu16veri776} dataset contains 49,360 images of 776 vehicles, of which the samples are obtained by 20 cameras on a circular road in a 1.0 Square Kilometers area for a short period of time (4:00 PM to 5:00 PM during the day), with each vehicle being captured by at least 2 and at most 18 cameras. 
VehicleID~\cite{Liu16CMC} includes 221,763 images about 26,267 vehicles, mainly containing both front and rear views.
For comprehensive evaluation of the vehicle Re-ID methods, VehicleID~\cite{Liu16CMC} divides the test set into 3 subsets, large, medium and small, according to the size of the vehicle images.
VehicleID~\cite{Liu16CMC} contains limited views (only two views, i.e., front view and rear view). In addition, the images in this dataset mainly contain less complex backgrounds, occlusions and illumination changes. 
VERI-Wild~\cite{Lou19veriwild} is collected in a 200 Square Kilometers suburban areas and contains 416,314 images of 40,671 vehicles taken by 174 traffic cameras.
The training set consists of 277,797 images (30,671 cars) and the testing set consists of 138,517 images (10,000 cars). Similarly, the testing set of VERI-Wild~\cite{Lou19veriwild} is divided into three subsets according to image size: large, medium, and small. The vehicle images in VERI-Wild~\cite{Lou19veriwild} mainly have little variability in views, mostly in front and rear views.

Although impressive results have been achieved on these datasets, the vehicle Re-ID problem is still
far from being addressed in the real-world scenarios.
First, these datasets contain only a limited number of scenarios and cameras.
The samples in VeRI-776~\cite{Liu16veri776}, VehicleID~\cite{Liu16CMC} and VERI-Wild~\cite{Lou19veriwild} are captured by 20, 12 and 174 cameras, respectively.
This is inconsistent with the real-life surveillance system in the smart city which contains tens of thousands cameras.
Second, the distribution of vehicle views is uneven, with most vehicles containing images of only the front and rear views and lacking side images.
Moreover, the number of cameras that each vehicle crosses is limited in the existing datasets, and thus it is difficult to evaluate the cross-scene retrieval capability of the models.

\section{VCNet: Viewpoint-conditioned Network}
In this work, to effectively combine the complementary information from different vehicle viewpoints, we propose a novel viewpoint-conditioned network (VCNet) for multi-query vehicle Re-ID. 
\subsection{Network Architecture}
\begin{figure}[t]
\begin{center}
\includegraphics[width=0.95\columnwidth]{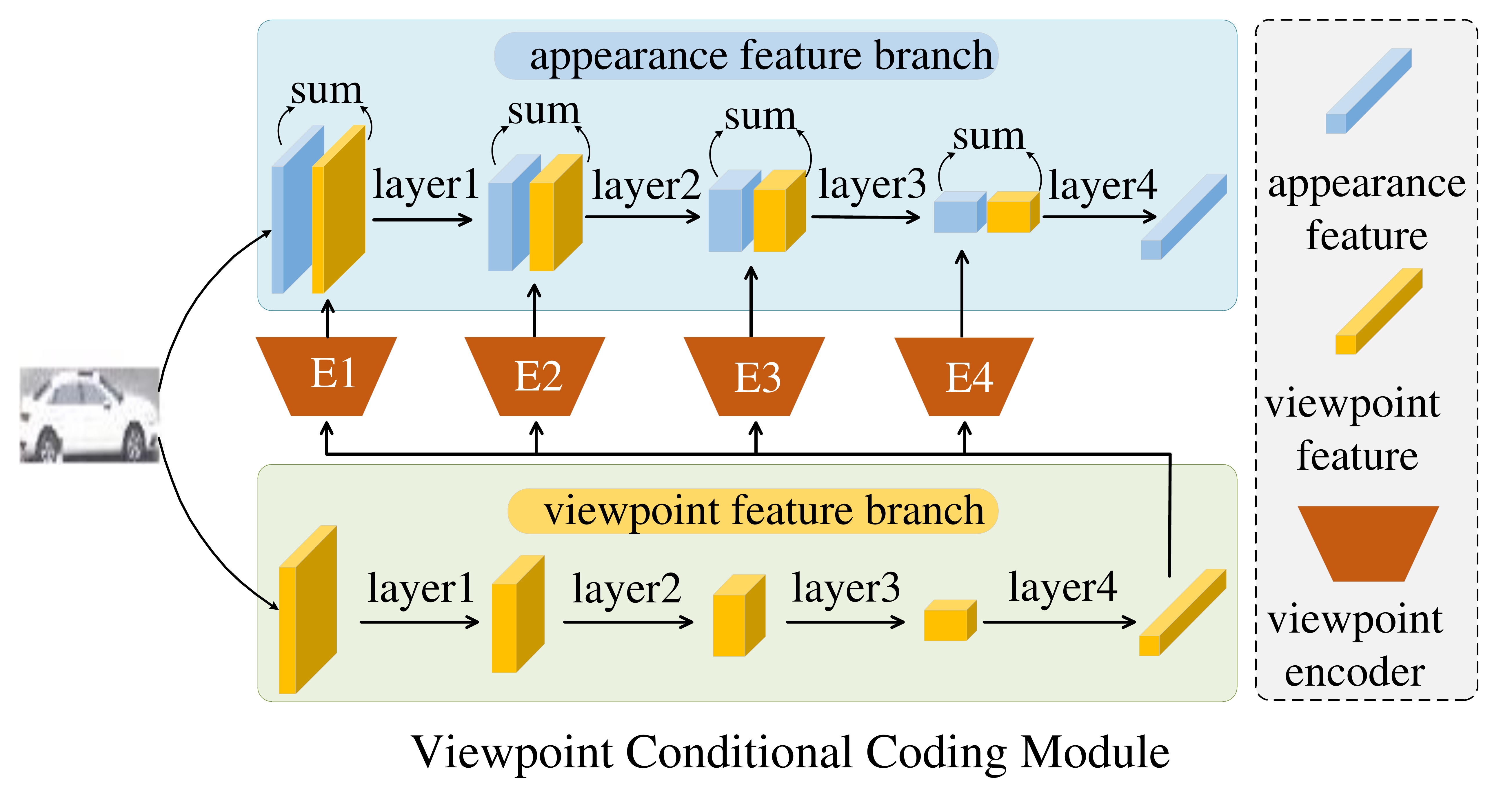} 
\end{center}
\caption{
The framework of the proposed VCC module. First, we learn the vehicle viewpoint features by the yellow branch below, then pass them through different deconvolution encoders (E1, E2, E3, and E4) to obtain viewpoint encoding features in different scales. These viewpoint encoding features are added to the vehicle appearance feature learning branch to learn vehicle detail features based on specific viewpoints.
}
\label{fig:vccnet}  
\end{figure}

Our VCNet includes two stages: multi-query inference as shown in Fig.~\ref{fig:test}.
First, we propose a viewpoint conditional coding (VCC) module to learn specific viewpoint information.
By encoding the vehicle's viewpoint features and embedding them into the learning process of vehicle detail features, it enforces the model to focus on the detail information under a specific viewpoint of the vehicle. As shown in Fig.~\ref{fig:vccnet}, we use the vehicle's viewpoint features as conditional encoding information, to fuse with the vehicle detail features obtained at each layer of the network.
It thus enables the model to focus on the vehicle viewpoint information while learning the discriminative features at that viewpoint.
%
%
%

To integrate the complementary information among different viewpoints of the vehicle, we propose a viewpoint-based adaptive fusion (VAF) module for multi-query inference. 
As shown in Fig.~\ref{fig:test}, we first assign the appearance feature weights of the query according to the similarity between the multi-query and gallery viewpoint features, then adaptively fuse the features of the multi-query according to the obtained weights, so as to take into account the complementarity and specificity between the different viewpoint features of the vehicle.
%
%
%

To handle the scenario of multi-query images with missing viewpoints, we further propose a cross-view feature recovery (CVFR) module to recover the missing appearance features.
CVFR module maximizes the common information between different viewpoints through comparative learning, and completes the reconstruction between different viewpoint features based on the common information. 

\begin{figure}[ht]
\begin{center}
\includegraphics[width=1.0\columnwidth]{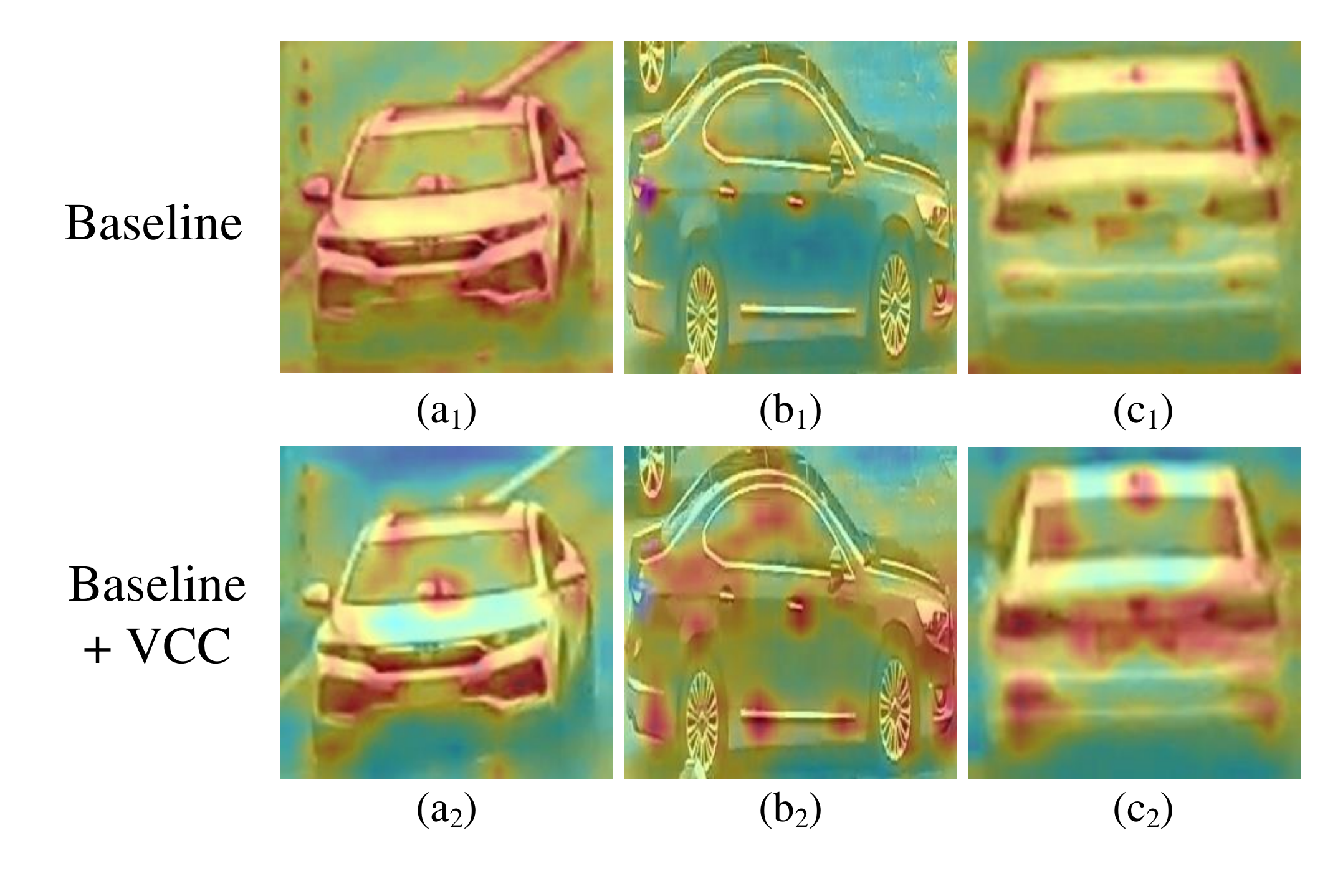} 
\end{center}
\caption{
Visualization of the feature maps of our VCC module comparing with the baseline. }

\label{fig:featureVisual}
\end{figure}

\subsection{Viewpoint Conditional Coding Module}
The large intra-class variability due to different viewpoints is a huge challenge for vehicle Re-ID.
Therefore, for the vehicle with different viewpoints, the network should focus on different detailed regions.
To make full use of the information of vehicle viewpoint, we propose a viewpoint conditional coding (VCC) module, as shown in Fig.~\ref{fig:vccnet}.
We introduce a two-stream network structure in the VCC module, the upper and lower branch is used to learn the appearance and viewpoint features of the vehicle respectively, and both branches use ResNet-50~\cite{he2016deep} as feature extractor.

First, different from Wang \textit{et al.}~\cite{wang17keypoint} which mark 8 viewpoints ($front$, $rear$, $left$, $left$ $front$, $left$ $rear$, $right$, $right$ $front$ and $right$ $rear$) on VeRI-776 dataset, we re-divide the 8 viewpoints into 3 viewpoint labels ($front$, $rear$ and $side$) to maximize the variation between different viewpoints for the training of viewpoint prediction.
To obtain a more robust and refined viewpoint features, on the basis of the training on the VeRI-776 dataset, we re-train the viewpoint prediction network on our MuRI dataset. 
To regress viewpoints, we use the cross-entropy loss as the supervision of the training of viewpoints as follows,
\begin{equation}
    \mathcal{L}_{view}=-\frac{1}{N}\sum_{i=1}^{N}log(p(v_i|x_i)),
\end{equation}
where $N$ represents the number of images in a training batch, $x_i$ denotes the input image, and $v_i$ denotes the viewpoint label.

Then, to enforce the network focus on more discriminative regions based on the viewpoint information, the learned viewpoint features are encoded to the vehicle appearance feature learning branch.
Here, we use different deconvolution functions as the viewpoint encoders to map the viewpoint features to the embedded features whose dimensions are same with the corresponding layers.
Next, we sum the appearance features and the embedded features and then send to the next layer of the network as viewpoint encoding information.
Finally, we can obtain the vehicle features which contain specific viewpoint information. The cross-entropy loss and triplet loss are used for the training of appearance features as follows,
\begin{equation}
\begin{split}
     \mathcal{L}_{appearance}=-\frac{1}{N}\sum_{i=1}^{N}log(p(y_i|x_i)) + \\
             \frac{1}{N}\sum_{i=1}^{N}(m + d(f_a^{i},f_p^{i}) - d(f_a^{i},f_n^{i})),
\end{split}
\end{equation}
where $y_i$ denotes the appearance label, $m$ denotes the margin, $d(\cdot)$ indicates the Euclidean distance, $f_a$, $f_p$ and $f_n$ denotes anchor, positive and negative appearance features respectively.

At last, the training loss of VCC module can be formulated as,
\begin{equation}
    \mathcal{L}_{vcc}= \mathcal{L}_{view} + \mathcal{L}_{appearance}.
\end{equation}

To demonstrate the effectiveness of VCC module, we visualize the features of the last layer, as shown in Fig.~\ref{fig:featureVisual}.
VCC module can reduce the
interference of background and enforce the model to better focus on the vehicles, comparing Fig.~\ref{fig:featureVisual} ($a_2$) with Fig.~\ref{fig:featureVisual} ($a_1$).
In addition, 
VCC module encourages the model to focus on the main clues for
classification and explores more discriminative regions, comparing Fig.~\ref{fig:featureVisual} ($b_2$) and ($c_2$) with Fig.~\ref{fig:featureVisual} ($b_1$) and ($c_1$). 

\subsection{Viewpoint-based Adaptive Fusion}
Although we have got a robust features that contains viewpoint information in VCC module, the limited information in a single query image significantly hinders the performance of vehicle Re-ID in the inference stage.
To integrate multi-viewpoint information of the vehicle in the inference stage and solve diverse viewpoint gaps between query and gallery, we propose a viewpoint-based adaptive fusion (VAF) module in the inference process, which adaptively fuses the generated viewpoint weights with appearance features.
%
%
%
%
The inference process is shown in Fig.~\ref{fig:test}.

First, we jointly use 3 vehicle images with different viewpoints in the query set and send them into the pre-trained VCC module to extract the appearance features and viewpoint features respectively.
To obtain the viewpoint similarity between 3 query images with gallery images, we calculate the features cosine distance between 3 query viewpoint features with gallery viewpoint features as follows,
\begin{equation}
\begin{split}
     s_i = \frac{{<f_{v}^{q_i},f_{v}^{g}>}}{||f_{v}^{q_i}||\times||f_{v}^{g}||},
\end{split}
\end{equation}
where $i=1,2,3$, $f_{v}^{q_i}$ and $f_{v}^{g}$ denotes query and gallery viewpoint features, $<x,y>$ indicates the inner product of $x$ and $y$.

Then, we can obtain the similarity weight set $\textbf{W} =$ \{ $w_i| i=1,2,3$  \} by computing the similarity of query and gallery viewpoint features using the concatenation and softmax function.
To adaptively fuse the viewpoint information in the appearance features, we multiply the multi-query appearance features $\textbf{F} =$ \{ $f_{a}^{q_i}| i=1,2,3$  \} with the similarity weight set $\textbf{W}$ to obtain the weighted appearance features $F{}' $ as follows,
\begin{equation}
\begin{split}
     \bar{F}  =\left \{ \bar{f_a^{q_1}}\times w_1, \bar{f_a^{q_1}}\times w_2, \bar{f_a^{q_1}}\times w_3\right \},
\end{split}
\end{equation}


%
To this end, the query appearance features with the similar viewpoint as the gallery image will be assigned a large weight.
If a query image from a random viewpoint is missing, the appearance feature will be recovered by the cross-view feature recovery (CVFR) module.
%
For the final recognition task, we perform a similarity calculation between the fused appearance features with the gallery appearance features and obtain the corresponding scores, which are summed to obtain the final recognition scores.

\subsection{Cross-view Feature Recovery Module}
In some scenarios, the query data might not contain some viewpoints of vehicle images. While our network accepts three query vehicle images as inputs, and thus can not handle such data with missing viewpoints. 
To solve this problem, referring to Lin \textit{et al.}~\cite{Lin21MV} in multi-view, we propose cross-view feature recovery (CVFR) module to recover the missing appearance features. 
To learn information-rich consistent representations, CVFR module maximizes the mutual information between different viewpoints by contrast learning.
To recover the missing appearance features, CVFR module minimizes the conditional entropy of different viewpoints by dual prediction.
For the sake of convenience, we assume that one viewpoint from $front$, $rear$, and $side$ is randomly missing, and the recovery process of the missing appearance features is as follows.
First, we send two known images from different viewpoints to the pre-trained VCC module and obtain the appearance features $x_1$ and $x_2$, respectively.
Then the latent representations $Z_1$, $Z_2$ are obtained after the respective encoders $E_1$, $E_2$, and the reconstructed features $x_1^{\prime}$, $x_2^{\prime}$ are obtained after the decoders $D_1$, $D_2$.
The reconstructed differences will be minimized by the mean squared error loss function as follows,
\begin{equation}
   \mathcal{L}_{mse}=\sum_{v=1}^{2}\sum_{t=1}^{m}||x_v^t-x_v^{\prime t}||^2, 
\end{equation}
where $x_v^t$ denotes the $t$-th sample of $x_v$.
To facilitate data recovery ability, contrastive learning is used to learn the common information between different viewpoints and to maximize the common information. The contrastive loss mathematical formula is as follows:
\begin{equation}
    \mathcal{L}_{cl}=-\sum_{t=1}^{m}(I(Z_1^t,Z_2^t)+\alpha(H(Z_1^t)+H(Z_2^t))),
\end{equation}
where $I$ denotes the mutual information, $H$ is the information entropy, and the parameter $\alpha$ is set as 9 to regularize the entropy in our experiments. From information theory,
information entropy is the average amount of information conveyed by an event. Hence a larger entropy $H(Z^i)$ denotes a more informative representation $Z^i$.
The viewpoint predictors $G_1$, $G_2$ are used to generate latent representations of the corresponding viewpoints, and the differences are generated by minimizing the loss function,
\begin{equation}
    \mathcal{L}_{pre}=\sum_{v=1}^{2}\sum_{t=1}^{m}||g_v^t-Z_{3-v}^t||^2, 
\end{equation}
where $v$ represents the number of available viewpoints, and we can obtain the missing appearance feature with random viewpoints from available ones.

To further narrow the differences between the generated and the original viewpoint features, we feed the latent representations generated by the viewpoint predictor into the corresponding decoders separately.
Then we obtain the generation of reconfiguration features $g_1^{\prime}$, $g_2^{\prime}$, which are constrained by mean squared error loss,
\begin{equation}
    \mathcal{L}_{mse}=\sum_{v=1}^{2}\sum_{t=1}^{m}||x_v^t-D_v(g_k^t)||^2, 
\end{equation}
where $D_v$ denotes Decoders.

\section{MuRI Dataset}
To evaluate the proposed VCNet on multi-query vehicle Re-ID, we propose a multi-views and unconstrained vehicle Re-ID dataset, MuRI, to integrate the complementary information among different viewpoints during inference.

\begin{figure}[t]
\begin{center}
\includegraphics[width=1.0\columnwidth]{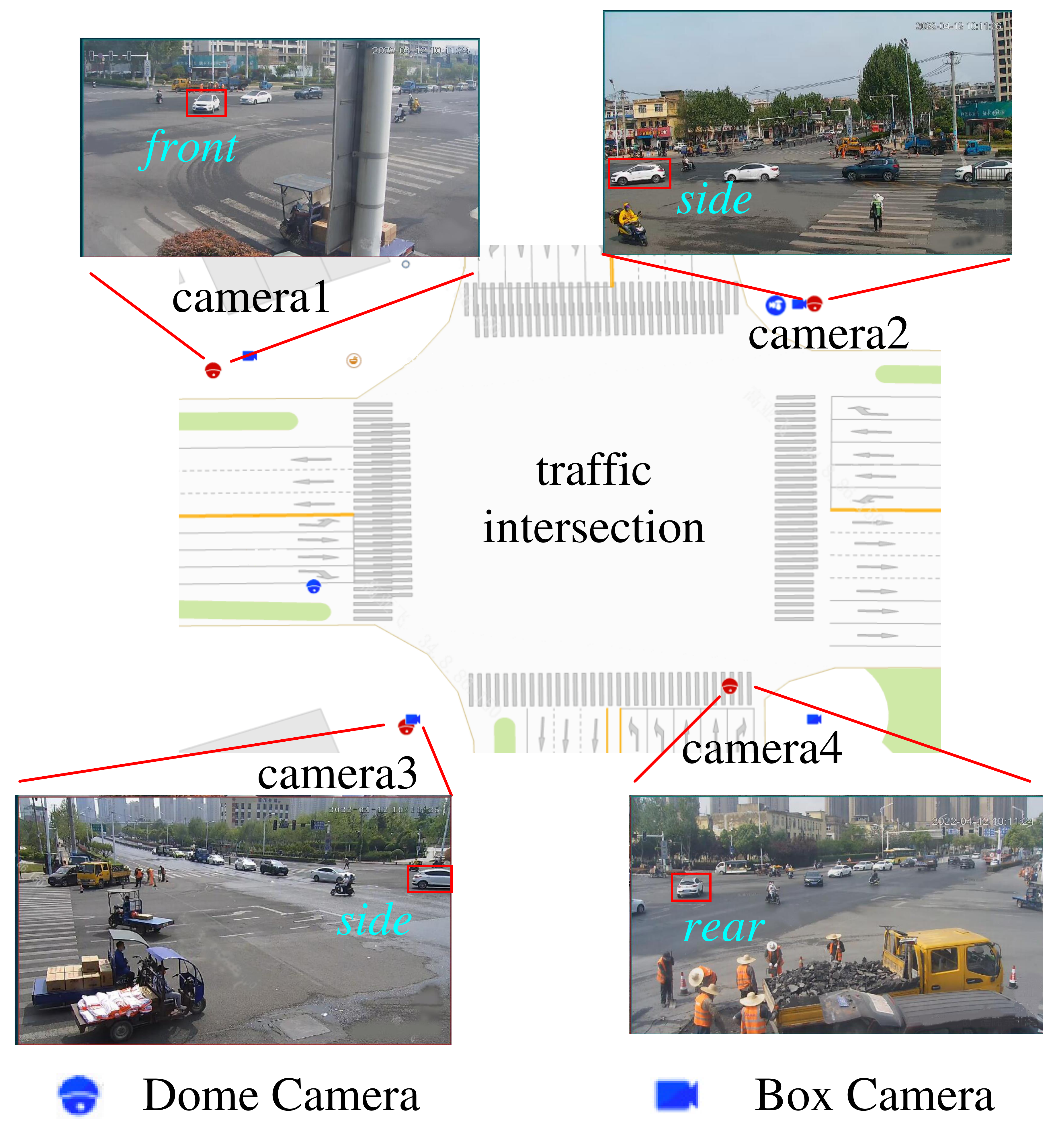}
\end{center}
\caption{
Illustration of data acquisition environment of the vehicle images obtained in the MuRI dataset. Vehicle images with diverse viewpoints are captured by the dome cameras at traffic intersection.
}
\label{fig:intersection}
\end{figure}

\noindent
\textbf{Data Acquisition.}
The MuRI dataset is collected in a large city with more than 1000 Squares Kilometers.
to obtain the vehicle images from more diverse cameras, we first search the corresponding vehicle images by license plate in the Public Security City Service Platform, which monitors tens of thousands of cameras in the city.
To ensure that each vehicle has rich viewpoint information, we obtain the vehicle images of different viewpoints at a traffic intersection, which has three or four surveillance cameras from different directions.
As shown in Fig.~\ref{fig:intersection},
to ensure the diverse viewpoint information, we choose the rotatable dome cameras from the Public Security City Service Platform as the shooting cameras.
For the videos captured from the platform, we generate the surrounding boxes by the tracking detection algorithm~\cite{yolov4}.
For effective evaluation, we automatically select the vehicle images in every 3 adjacent frames, followed by manual checking to avoid data redundancy.%
The time span of vehicle appearance in the data set is about half the year, and the vehicle resolution is variable due to the varying distances between cameras and vehicles of interest.
%

\noindent   
\textbf{Dataset Description.}
The MuRI dataset contains 200 identities in five viewpoints ($front$, $side$ $front$, $side$, $rear$, and $side$ $rear$) with diverse resolution and illumination conditions.
Due to the similar appearance between $front$ and $side$ $front$, as well as $rear$, and $side$ $rear$, we merge the five viewpoints into three in this paper, i.e., $front$, $side$ and $rear$.
For effective evaluation, we automatically select the vehicle images from the traffic intersection in every 3 adjacent images, followed by manual checking to avoid data redundancy.
Together with the images collected by the Public Security City Service Platform, our MuRI forms 23637 vehicle images from 6142 cameras in total.
We select 150 identities for training, and 50 identities for testing/inference. 
In the inference stage, we use the entire testing set as gallery set, while randomly selecting 3 records from different viewpoints as multi query images.

\begin{figure}[ht]
\begin{center}
\includegraphics[width=0.99\columnwidth]{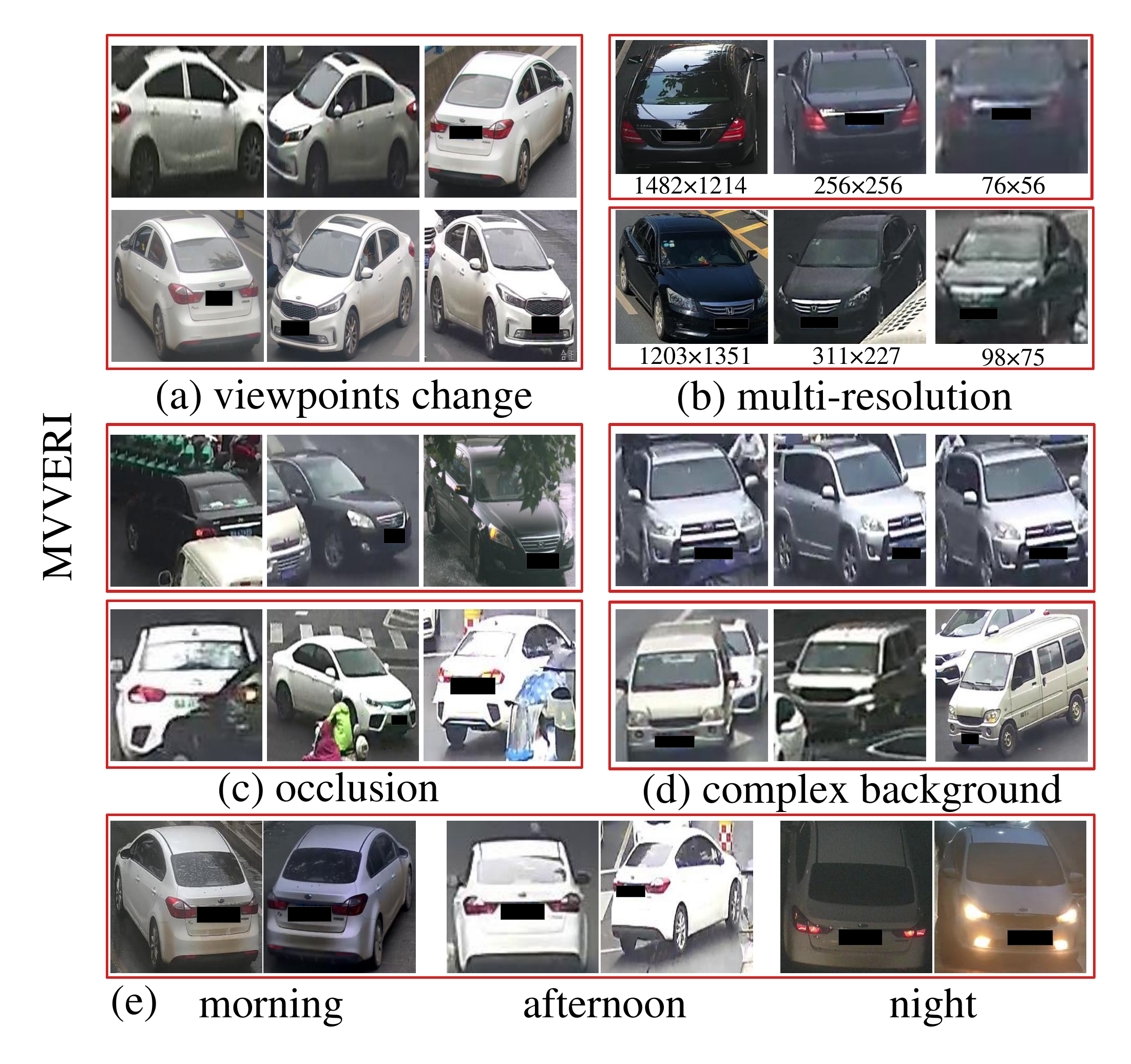}
\end{center}
\caption{
Challenges of MuRI dataset. The images in the common red box indicate the same vehicle ID.
}
\label{fig:datasetchallengs}
\end{figure}

\noindent   
\textbf{Dataset Challenges.}
Our MuRI mainly contains five different challenges as shown in Fig.~\ref{fig:datasetchallengs}.
First, our dataset provides comprehensive five viewpoints, including $front$ ($side$ $front$), $side$ and $rear$ ($side$ $rear$) for each vehicle, which produce huge intra-class for Vehicle Re-ID.
Then, due to the varying distances between cameras and vehicles of interest, the vehicle images present different resolutions as shown in Fig.~\ref{fig:datasetchallengs} (b).
The poor detailed information in the low resolution as well as the appearance gap between different resolutions further bring huge challenge for veichle Re-ID.
Moreover, MuRI dataset is collected in a large city surveillance system spanned more than 1000 $km^2$, and the urban environment is very complex.
To this end, MuRI contains many vehicle images with occlusion and complex background as shown in Fig.~\ref{fig:datasetchallengs} (c, d), which brings severe challenges for vehicle Re-ID.
Finally, the vehicle images in MuRI dataset are collected in a long time span with more than half a year, which provides large number of cross-time vehicle data with different illuminations, such as morning, afternoon, and evening as shown in Fig.~\ref{fig:datasetchallengs} (e).
The large illumination changes results in huge difference in vehicle appearance. Furthermore, the strong lighting from the headlights and the taillights during the night brings additional challenge for vehicle Re-ID.

\begin{table}[h]
\caption{
Comparisons among VehicleID, VeRI-776, VERI-Wild, and the created MuRI datasets for vehicle ReID.
} 

\resizebox{\linewidth}{!}{
\begin{tabular}{|c|c|c|c|c|}
\hline
Dataset            & VehicleID & VeRI-776 & VERI-Wild & MuRI \\ \hline
Images             & 221,763   & 49,360   & 416,314   & 23,637 \\ \hline
Identities         & 26,267    & 776      & 40,671    & 200    \\ \hline
Cameras            & 12        & 20       & 174       & 6142   \\ \hline
Viewpoints/id         & 2.0     & 4.2        & 3.4    & 5.0     \\ \hline
Cross-resolution   & $\times$         & $\times$       & $\times$         & $\surd$      \\ \hline
Occlusion          & $\times$         & $\times$        & $\surd$         & $\surd$      \\ \hline
Complex Background & $\times$         & $\times$        & $\surd$         & $\surd$      \\ \hline
Morning            & $\surd$         & $\times$        & $\surd$         & $\surd$      \\ \hline
Afternoon          & $\surd$         & $\surd$        & $\surd$         & $\surd$      \\ \hline
Night              & $\times$         & $\times$        & $\surd$         & $\surd$      \\ 
\hline
\end{tabular}}
\label{fig:data}
\end{table}

\begin{figure}[ht]
\begin{center}
\includegraphics[width=0.99\columnwidth]{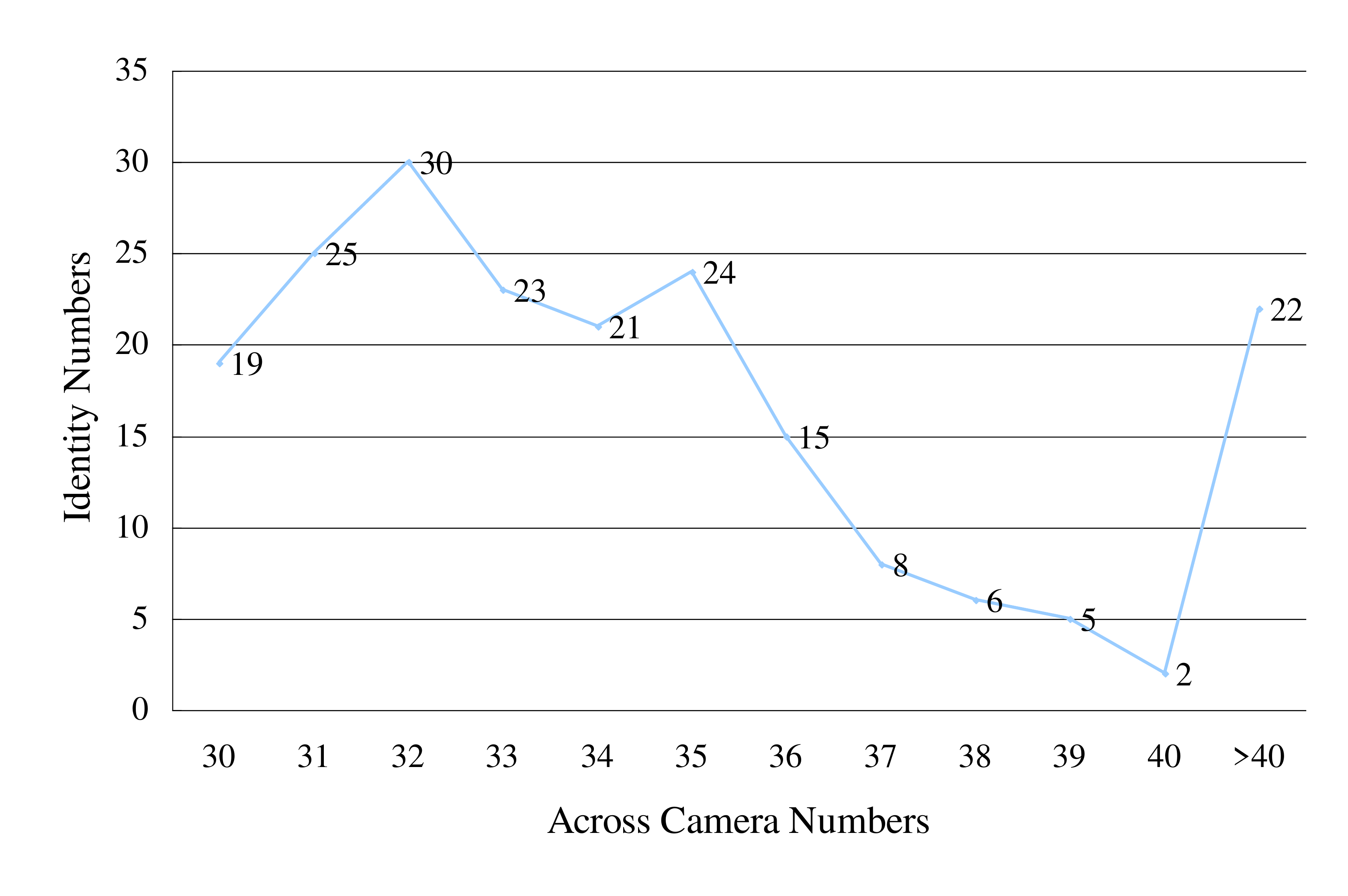}
\end{center}
\caption{
Distribution of the number of identities across the number of cameras.
}
\label{fig:datasetvisual}
\end{figure}

\noindent 
\textbf{Dataset Characteristics.}
Compared with existing prevalent Re-ID datasets as
shown in Table~\ref{fig:data}, in calculating the average number of viewpoints for each id in the dataset, we divided the vehicle viewpoints into five categories according to $front$, $side$ $front$, $side$, $side$ $rear$, and $rear$. MuRI has the following major advantages.
%
\begin{enumerate}
    \item
    {\bf Numerous cameras with wide area}. MuRI contains 200 vehicle IDs captured by 6142 cameras from a real-life transportation surveillance system covering over 1000 $km^2$ urban area.
    \item
    {\bf Comprehensive viewpoints of each ID}.
MuRI provides comprehensive five viewpoints, including $front$ ($side$ $front$), $side$ and $rear$ ($side$ $rear$) for each vehicle, which provides a more realistic and challenging scenario for vehicle Re-ID.
    \item
    {\bf Large number of cameras crossed by each ID}.
    Each vehicle in MuRI crosses 34.6 cameras in average, varying from 30 to 50 cameras, as shown in Fig.~\ref{fig:datasetvisual}.
\end{enumerate}

\begin{figure}[ht]
\begin{center}
\includegraphics[width=0.99\columnwidth]{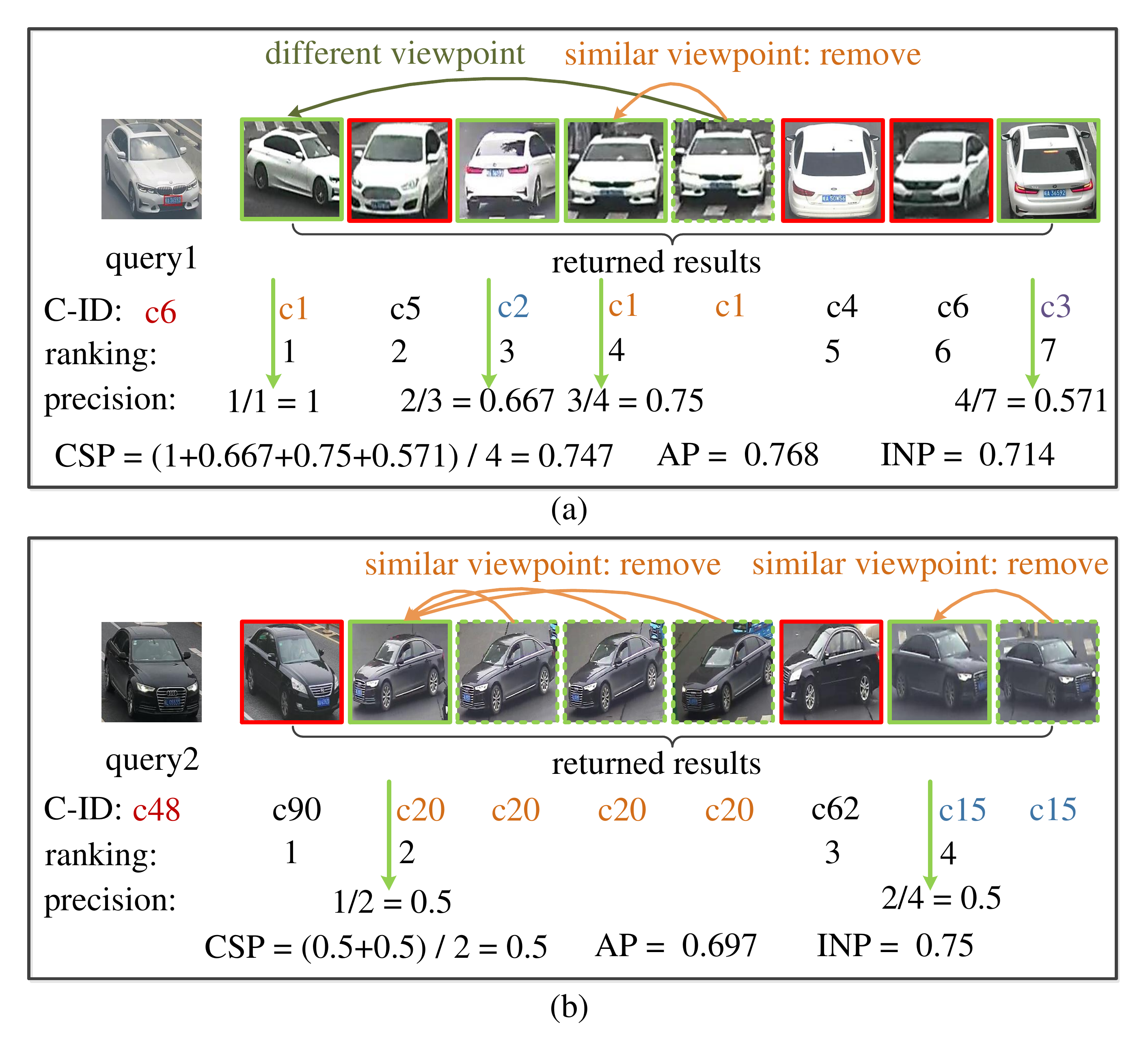}
\end{center}
\caption{
Illustration of the calculation process of CSP metric. C-ID denotes the camera id, true matching and false matching are bounded in green and red boxes,  respectively. For positive samples retrieved from the same camera, the CSP metric removes them during the calculation when their viewpoints are similar.
}
\label{fig:zhibiao}
\end{figure}

\section{mCSP Metric}
One problem with existing metrics is the lack of consideration of cross-scene scenarios. 
To this end, a new metric, named mean Cross-scene Precision (mCSP) is proposed in this paper to ensure the cross-scene retrieval capability of the network.
The main idea of mCSP can be summarized as follows: if there exist positive samples with similar viewpoints from the same camera, we consider them as the same scene and remove them from the ranked list. 
Given a ranked list, we use $TP$ to denote the number of positive samples retrieved, $f_{c}^{i}$ and $f_{c}^{j}$ denotes the viewpoint feature of the two positive sample image retrieved under the same camera $c$. When their Euclidean distance $d(f_{c}^{i},f_{c}^{j})$ is smaller than a threshold hyperparameter $\varepsilon$, we consider that the viewpoints of $f_{c}^{j}$ and $f_{c}^{i}$ is similar. 
We use $SC$ to denote the number of samples with the similar viewpoint under the same camera ID in $TP$, and $FP$ denotes the number of positive samples with prediction errors, mCSP can be expressed in the following form,
\begin{equation}
    mCSP = \frac{\sum_{i=0}^{N_{cs}}\frac{TP-SC}{TP-SC+FP}}{N_{cs}},
\end{equation}
where $N_{cs}$ denotes the captured target images from positive samples with different cameras. We visualize the calculation process of CSP, as shown in Fig.~\ref{fig:zhibiao}.
%
As shown in Fig.~\ref{fig:zhibiao} (a) and (b), for the positive samples detected under the same camera, CSP removes the ones with similar viewpoints.
Positive samples with similar viewpoints under the same camera tend to be more easily identified, which results in virtually high scores in the existing metrics such as AP and INP, comparing Fig.~\ref{fig:zhibiao} (b) to (a).
By contrast, 
the proposed CSP metric can better reflect the ability of cross-camera retrieval, as shown in Fig.~\ref{fig:zhibiao} (b). 

\section{Experiments}
\subsection{Experiments Setup}
\noindent
{\bf Train.}
We use ResNet-50~\cite{he2016deep} pre-trained on ImageNet~\cite{deng2009imagenet} as our backbone. The model is trained for 80 epochs with the SGD optimizer.
We warm up the learning rate to 5e-2 in the first 5 epochs and the backbone is frozen in the warm-up step. The learning rate of 5e-2 is kept until the 60th, drops to 5e-3 in the 60th epoch, and drops to 5e-4 in the 75th epoch for faster convergence. We first pad 10 pixels on the image border, and then randomly crop it to 256×256. We also augment the data with random erasing. Further, we add a Batch Normalization layer after the global feature. A fully connected layer is added to map the global feature to the ID classification score. The batch size is 36 in the MuRI dataset.

\begin{figure}[t]
\begin{center}
\includegraphics[width=1\columnwidth]{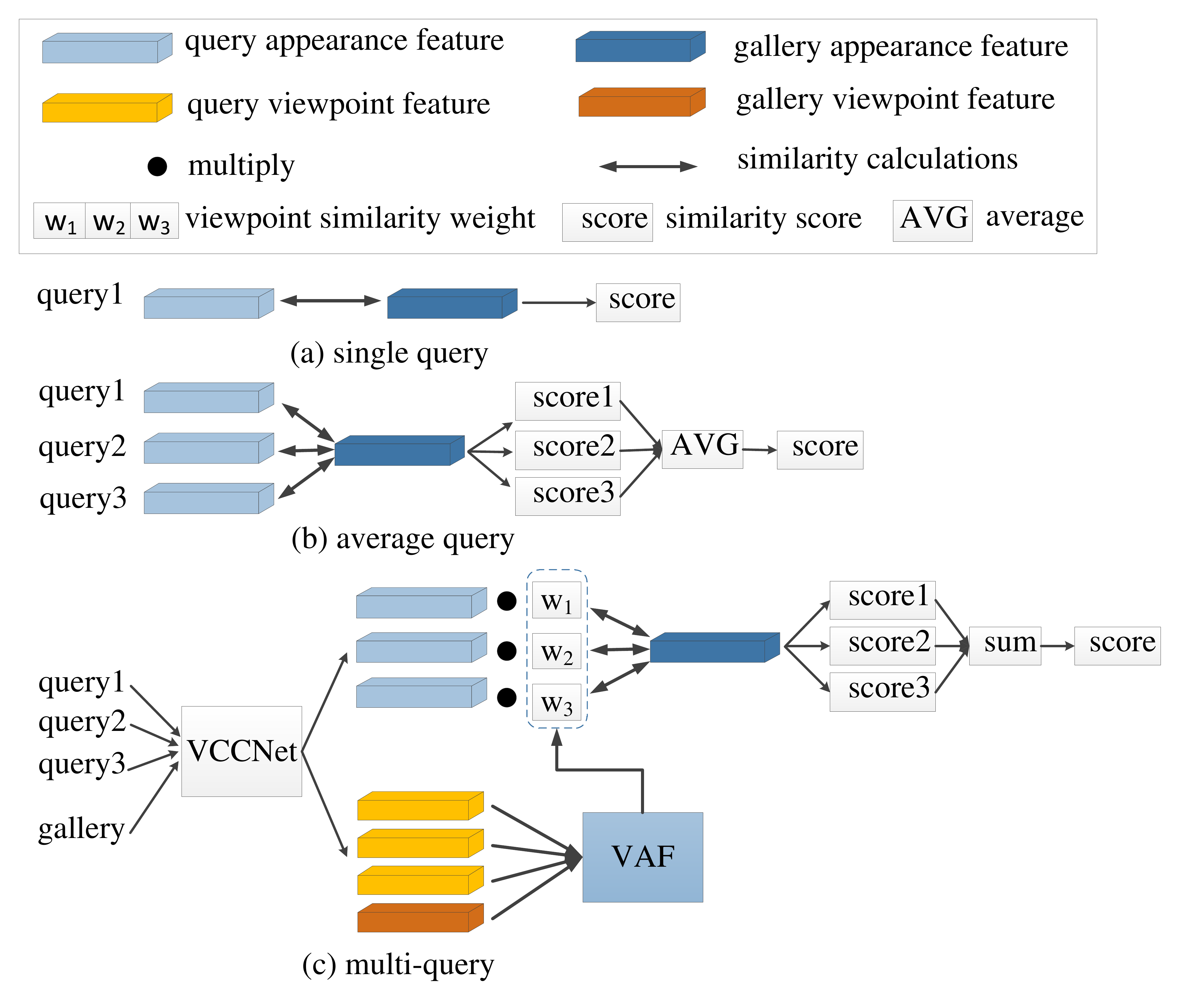} 
\end{center}
\caption{
Diagrams of different inference settings.
}
\label{fig:querysetting}
\end{figure}

\noindent
{\bf Inference.}
In our inference process, we evaluate the methods in three inference ways, including single query, average query and multi-query. 
As demonstrated in Fig.~\ref{fig:querysetting} (a), single inference directly calculate the cosine distance between each query and the gallery set, which ignores the multi-view information during the inference. 
When facing the multiple queries, the intuitive inference way is the average inference, which computes the average value of multiple query features, as shown in Fig.~\ref{fig:querysetting} (b).
However, simply averaging the query features can not effectively use the different viewpoint information of the vehicle.
%
To adaptively utilize the complementary information in the multiple queries from different cameras with diverse viewpoints combinations, we propose the multi-inference for the proposed VCNet, as shown in Fig.~\ref{fig:querysetting} (c).
Specifically, it generates viewpoint weights by the similarity between multiple queries and gallery view features, then fuses the generated viewpoint weights with appearance features.
\begin{table*}[ht] 
\centering
\caption{
Performance comparisons on MuRI benchmark. 
}
\resizebox{\linewidth}{!}{
\begin{tabular}{c|c|c|ccccccc}
\hline
Method                       & Venue  & Inference way & Rank1 & Rank5 & Rank10 & mAP   & mINP  & mCGM  & mCSP  \\ \hline
\multirow{2}{*}{DMML~\cite{chen2019dmml}}  &  \multirow{2}{*}{ICCV 2019}         & single query  & 0.644 & 0.767 & 0.814  & 0.362 & 0.060 & 0.175 & 0.198 \\
                             &  & average query & 0.766 & 0.884 & 0.927  & 0.524 & 0.088 & 0.281 & 0.272 \\ \hline
\multirow{2}{*}{RECT\_Net~\cite{zhu20rectnet}}  &  \multirow{2}{*}{CVPR 2020}    & single query  & 0.729 & 0.811 & 0.859  & 0.415 & 0.074 & 0.212 & 0.225 \\
                            &   & average query & 0.806 & 0.924 & 0.956  & 0.602 & 0.108 & 0.316 & 0.330 \\ \hline
\multirow{2}{*}{GRF~\cite{liu20grf}}   & \multirow{2}{*}{TIP 2020}  & single query  & 0.683 & 0.806 & 0.841  & 0.398 & 0.070 & 0.188 & 0.214 \\
                             &  & average query & 0.786 & 0.906 & 0.942  & 0.565 & 0.101 & 0.302 & 0.314 \\ \hline
\multirow{2}{*}{CAL~\cite{rao2021cal}}   &     \multirow{2}{*}{ICCV 2021}  & single query  & 0.754 & 0.842 & 0.890  & 0.441 & 0.082 & 0.228 & 0.266 \\
                             &  & average query & 0.816 & 0.960 & 0.980  & 0.645 & 0.138 & 0.360 & 0.359 \\ \hline
VCNet     & Ours  & multi-query  &\textbf{0.843}    &\textbf{0.962} & \textbf{0.980}  & \textbf{0.677} & \textbf{0.145} & \textbf{0.400} & \textbf{0.393} \\\hline
\end{tabular}}
\label{tab:sota2}
\end{table*}
%

\begin{figure}[t]
\begin{center}
\includegraphics[width=1.00\columnwidth]{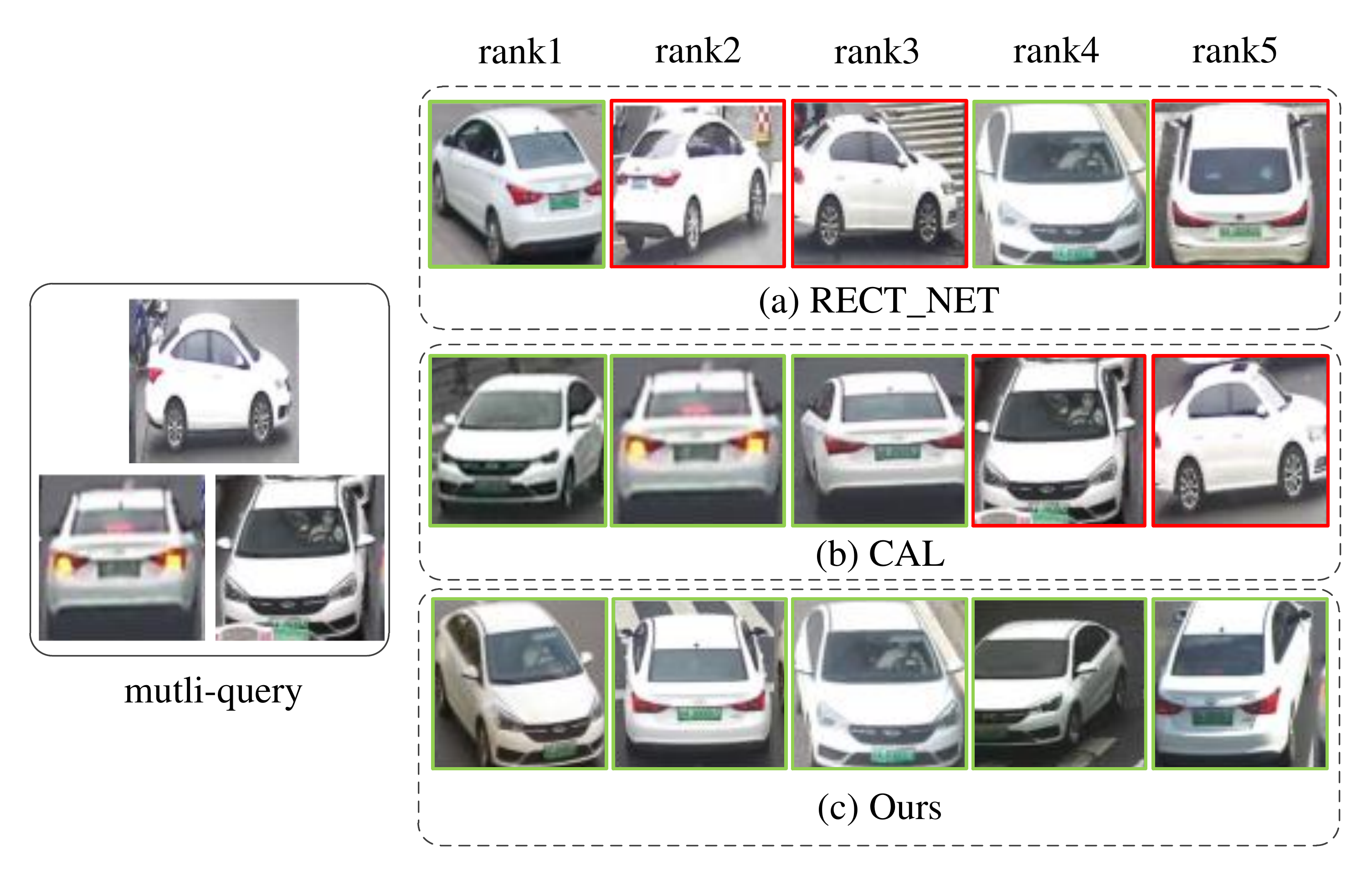}
\end{center}
\caption{
The top five results of different methods for multi-query inference.
}
\label{fig:comparion}
\end{figure}

\subsection{Experimental Results}
\noindent
\textbf{Comparison with the state-of-the-arts.}
To verify the effectiveness of the proposed VCNet with the multi-query setting, we compare four state-of-the-art vehicle Re-ID methods on the collected MuRI dataset.
Specifically, we construct the multi-query setting with the number of query $N_Q$ = 3, which including three different viewpoints $front$, $rear$ and $side$ respectively.
We evaluate the state-of-the-art methods in both single and average inferences for comparison. 
As shown in Table~\ref{tab:sota2}, 
all the state-of-the-art methods achieve significant improvement in the average inference compared to the single inference, which evidences that using multiple queries can better incorporate the complementary information among the images.
By progressively embedding viewpoint features to
appearance feature learning via the viewpoint conditional coding (VCC), and integrating the complementary information among different viewpoints via the viewpoint-based adaptive fusion (VAF), our VCNet with multi-inference achieves superior performance compared to the state-of-the-art methods.
This validates the effectiveness of the proposed VCNet while handling the multi-query inference for vehicle Re-ID.
Fig.~\ref{fig:comparion} shows the corresponding ranking results of multiple queries from different viewpoints, which further evidences the promising performance of our method while handling the challenging cross-scene retrieval problem compared to other methods.
\begin{table}[t]
\caption{
\textbf{}Ablation study of VCNet when the number of query $N_Q$ = 3 with one viewpoint random missing.
}
\resizebox{\linewidth}{!}{
\begin{tabular}{c|ccccc}
\hline
Settings      & \multicolumn{1}{c}{Rank1}  & \multicolumn{1}{c}{mAP} & \multicolumn{1}{c}{mINP} & \multicolumn{1}{c}{mCGM} & \multicolumn{1}{c}{mCSP} \\ \hline
Baseline      & 0.774            & 0.506          & 0.104        & 0.310          & 0.307          \\ \hline
+VCC          & 0.803        & 0.531           & 0.121         & 0.332          & 0.340            \\ \hline
+VCC+VAF      & 0.820       & 0.554        & 0.125     & 0.344          & 0.360          \\ \hline
\begin{tabular}[c]{@{}c@{}}+VCC+VAF\\ +CVFR\end{tabular}
  & 0.832    & 0.565        & 0.130      & 0.358      &  0.371        \\ \hline
\end{tabular}}
\label{tab:ablation}
\end{table}

\noindent
\textbf{Ablation study of VCNet.}       
To verify the effective contribution of the components in our model, we implement the ablation study on the viewpoint conditional coding (VCC) module, viewpoint-based adaptive fusion (VAF) module, and cross-view feature recovery (CVFR) module on our MuRI dataset, as shown in Table~\ref{tab:ablation}.
We employ ResNet-50~\cite{he2016deep} in a single query fashion as our baseline to extract vehicle appearance features.
Note that introducing VCC significantly boosts the baseline, which evidences the effectiveness of the proposed VCC module which can integrate the complementary information among different viewpoints of the vehicle.
VAF consistently brings a significant improvement on all the metrics by adaptively fusing the generated viewpoint weights with appearance features.
At last, CVFR further enhances the performance by recovering features of the missing viewpoint.

\begin{table}[t]
\centering
\caption{
Evaluation of VAF when the number of queries $N_Q$=3.
}
\resizebox{\linewidth}{!}{
\begin{tabular}{ccccccc} \hline
& \multicolumn{1}{c|}{Methods}     & Rank1   & mAP    & mINP  & mCGM & mCSP\\ \hline
& \multicolumn{1}{c|}{RECT\_NET~\cite{zhu20rectnet}}
& 0.806          & 0.602      & 0.108      & 0.316      & 0.330    \\ 
& \multicolumn{1}{c|}{RECT\_NET + VAF}
& 0.820         & 0.620      & 0.114      & 0.349      & 0.338     \\ \hline

& \multicolumn{1}{c|}{CAL~\cite{rao2021cal}} 
& 0.816         & 0.645      & 0.138      & 0.360      & 0.359     \\ 
& \multicolumn{1}{c|}{CAL + VAF}
& 0.838         & 0.667      & 0.142      & 0.372      & 0.380     \\ \hline


&  \multicolumn{1}{c|}{VCC (Ours)}
& 0.826         & 0.659      & 0.141      & 0.374      & 0.370         \\ 
&   \multicolumn{1}{c|}{VCC + VAF}  
& \textbf{0.843}         & \textbf{0.677}      & \textbf{0.145}      & \textbf{0.400}      & \textbf{0.393}       \\ \hline
\end{tabular}}
\label{tab:vaf}
\end{table}

\noindent
{\bf Evaluation on VAF.}
To further demonstrate the effectiveness and applicability of viewpoint-based adaptive fusion (VAF) module, we plugin VAF into two state-of-the-art methods with the number of query $N_Q$ = 3 with three different viewpoints ($front$, $rear$ and $side$) in the query set.
To obtain the viewpoint features for the VAF module, we use a pre-trained viewpoint prediction network for all the other methods.
In vehicle re-identification, the main challenge is the intra-class variability and inter-class similarity problem due to the difference in vehicle viewpoints. We can use the images from different views of the vehicle during the inference through multi-query. VAF can assign weights adaptively according to the similarity between the viewpoints of vehicles in query and gallery. More similarity between the query and gallery viewpoints, the greater the weight when retrieving, which reduce the difficulty of identifying positive samples.
As shown in Table~\ref{tab:vaf}, after integrating the proposed VAF into RECT\_NET~\cite{zhu20rectnet} and CAL~\cite{rao2021cal}, it brings a large margin improvement over the original methods by fusing the generated viewpoint weights with their appearance features.
This verifies that VAF can better integrate the complementary information among different viewpoints.

\begin{table}[t] 
\centering
\caption{
Evaluation on CVFR when the number of query $N_Q=3$ with one random missing. 
}
\begin{tabular}{l|ccccc}
\hline
Inference way      & Rank1  & mAP  & mINP  &mCGM & mCSP  \\ \hline
(a) single       & 0.715  & 0.426 & 0.087 &0.256 & 0.272 \\ 
(b) average      & 0.801  & 0.535 & 0.122 &0.336 & 0.342 \\  
(c) CVFR + average & 0.814  & 0.544 & 0.126 &0.345 & 0.351 \\  
(d) multi (2 views)   & 0.820  & 0.554 & 0.127 &0.344 & 0.360 \\  
(e) CVFR + multi   & 0.832  & 0.565 & 0.130 &0.358 & 0.371 \\ \hline
\end{tabular}
\label{tab:cvfr}
\end{table}

\noindent
\textbf{Evaluation on CVFR.}
To handle the scenario with viewpoint missing, we propose the cross-view feature recovery (CVFR) module to recover the appearance features of the missing viewpoints. 
To validate the compatibility of our proposed multi-query inference method with viewpoint missing, we evaluate our method with different viewpoint missing cases as shown in Table~\ref{tab:cvfr}.
The average query as shown in Table~\ref{tab:cvfr} (b) makes a significant improvement by combining two viewpoints, compared to the single inference in Table~\ref{tab:cvfr} (a).
By recovering the missing appearance features via the proposed CVFR module, Table~\ref{tab:cvfr} (c) brings further improvement. 
Our proposed multi-query inference can further boost the performance with only two existing viewpoints, as shown in Table~\ref{tab:cvfr} (d), which strongly verifies the effectiveness of the proposed multi-query inference.
Finally, the multi-query inference together with recovering the missing appearance features via the proposed CVFR module achieves the best performance, as shown in Table~\ref{tab:cvfr} (e), which indicates the necessity of the supplementary information in multi-query for vehicle Re-ID. 

\begin{figure}[t]
\begin{center}
\includegraphics[width=1.0\columnwidth]{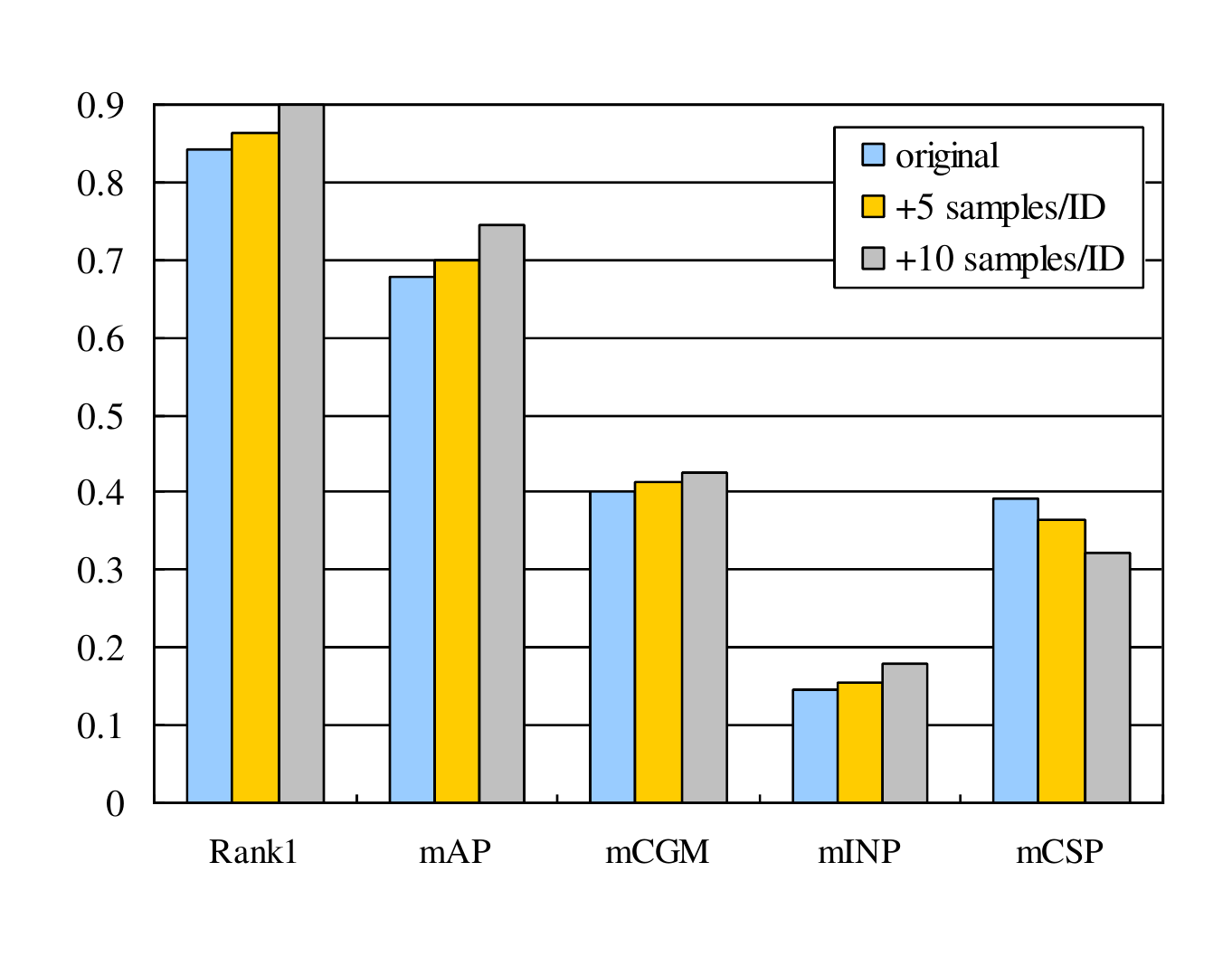}
\end{center}
\caption{
 The changes of each metric on the modified gallery sets in MuRI.
}
\label{fig:mcsp}
\end{figure}

\noindent   
\textbf{Evaluation on mCSP.}
To evaluate the capability of cross-scene retrieval of the Re-ID methods, we reconstruct by adding 5 and 10 images of the same viewpoint under the same camera for each vehicle in the original gallery set.
To make a fair comparison, we further delete 5 and 10 images of that vehicle under different cameras to ensure the total number of images of each vehicle remain unchanged. 
Fig.~\ref{fig:mcsp} shows the comparison of the proposed mCSP comparing with existing metrics on both the original and modified gallery sets.
%
%
By adding the images with the same viewpoint under the same camera, all the Rank1, mAP, and mINP increase in the modified galleries due to more easy matching samples.
This is irrational in realistic Re-ID where the capability of matching more positive vehicle images across more diverse scenes is even crucial.
By contrast, the proposed mCSP declines with the images with the same viewpoint under the same camera increase, since the positive samples recognized under different cameras become less. The mCSP only focuses on retrieving positive sample images from different cameras in the gallery set and images from different views in the same camera, which can more realistically reflect the retrieval accuracy across cameras in the realistic Re-ID.

\begin{figure}[t]
\begin{center}
\includegraphics[width=1\columnwidth]{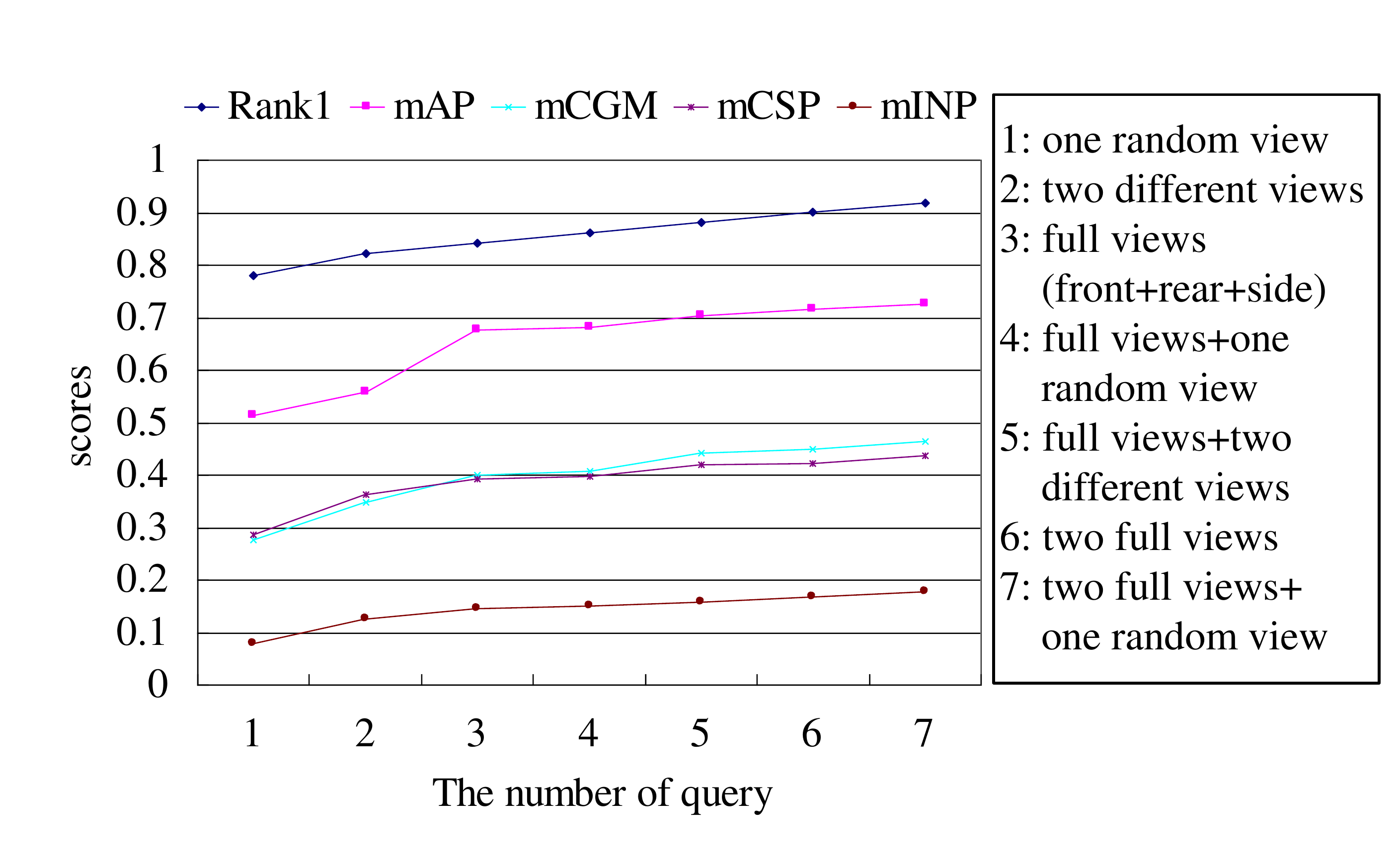} 
\end{center}
\caption{
Evaluation of the number of queries on MuRI. 
}
\label{fig:querynumber}
\end{figure}

\noindent
\textbf{Evaluation on the number of query}
Fig.~\ref{fig:querynumber} evaluates the multi-query inference with the different number of queries. We can see that as the number of queries increases, the performance of the multi-query consistently improves, benefiting from the richer information in the multiple diverse images about the vehicle.
Note that the significant improvement in each metric is achieved by increasing from 1 random view to 2 different views, and then to 3 full views ($front + back + side$). When the number of queries continues to increase from 3 to 7, the performance of each metric consistently increases, but with a slightly slower increase. This indicates that the larger diversity in viewpoints between queries, the better improvement in multi-query inference, since more complementary information between different viewpoints of the vehicle.
\section{Conclusion}
In this paper, we first launch the multi-query vehicle Re-ID task which leverages multiple queries to overcome the viewpoint limitation of a single one, and propose a viewpoint-conditioned network (VCNet) for multi-query vehicle Re-ID. 
First, we propose a viewpoint conditional coding (VCC) module
in the training process to learn specific viewpoint information.
Then, we propose a viewpoint-based adaptive fusion (VAF) module to integrate the complementary information among different viewpoints in the inference process.
To handle the scenario when query images from random viewpoints, we propose  the cross-view feature recovery (CVFR) module to recover the missing appearance feature.
Finally, a new metric (mCSP) and a new dataset (MuRI) are proposed to measure the ability of cross-scene recognition and conduct multi-query evaluation experiments respectively.
Comprehensive experiments demonstrate the necessity of the multi-query inference and the effectiveness of the proposed VCNet.
This work provides new research direction for vehicle Re-ID and related areas.

{\small
\bibliographystyle{ieeetr}
\bibliography{egbib}
}
\vfill

\end{document}